\documentclass{article}

     \PassOptionsToPackage{square,sort,comma,numbers}{natbib}

     \usepackage[preprint]{neurips_2021}

\usepackage[utf8]{inputenc} 
\usepackage[T1]{fontenc}    
\usepackage{hyperref}       
\usepackage{url}            
\usepackage{booktabs}       
\usepackage{amsfonts}       
\usepackage{nicefrac}       
\usepackage{microtype}      
\usepackage{xcolor}         
\usepackage{graphicx}
\usepackage{float}
\usepackage{url}
\usepackage{amsmath}
\usepackage{comment}
\usepackage{diagbox}
\usepackage{makecell}
\usepackage{algorithm}  
\usepackage{algorithmic} 
\usepackage{diagbox}

\usepackage{amsmath,amsfonts,bm}

\def\eqref#1{equation~\ref{#1}}

\def\1{\bm{1}}

\DeclareMathAlphabet{\mathsfit}{\encodingdefault}{\sfdefault}{m}{sl}
\SetMathAlphabet{\mathsfit}{bold}{\encodingdefault}{\sfdefault}{bx}{n}

\usepackage{color}
\usepackage{amsmath}
\usepackage{sidecap}

\title{What can we learn from misclassified ImageNet images?}

\author{%
  Shixian Wen\\
  Department of Computer Science\\
  University of Southern California\\
  Los Angeles, CA 90089 \\
  \texttt{shixianw@usc.edu} \\
     \And
    Amanda Sofie Rios \\
  Neuroscience Program\\
  University of Southern California\\
  Los Angeles, CA 90089 \\
  \texttt{amandari@usc.edu } \\
   \And
       Kiran Lekkala \\
  Department of Computer Science\\
  University of Southern California\\
  Los Angeles, CA 90089 \\
  \texttt{klekkala@usc.edu } \\
   \And
   Laurent Itti \\
  Department of Computer Science\\
  University of Southern California\\
  Los Angeles, CA 90089 \\
  \texttt{Itti@usc.edu} \\
}

\begin{document}

\maketitle

\begin{abstract}
Understanding the patterns of misclassified ImageNet images is particularly important, as it could guide us to design deep neural networks (DNN) that generalize better. However, the richness of ImageNet imposes difficulties for researchers to visually find any useful patterns of misclassification.  Here, to help find these patterns, we propose "Superclassing ImageNet dataset". It is a subset of ImageNet which consists of 10 superclasses, each containing 7-116 related subclasses (e.g., 52 bird types, 116 dog types).  By training neural networks on this dataset, we found that: (i) Misclassifications are rarely across superclasses, but mainly among subclasses within a superclass. (ii) Ensemble networks trained each only on subclasses of a given superclass perform better than the same network trained on all subclasses of all superclasses. Hence, we propose a two-stage Super-Sub framework, and demonstrate that: (i) The framework improves overall classification performance by 3.3\%, by first inferring a superclass using a generalist superclass-level network, and then using a specialized network for final subclass-level classification. (ii) Although the total parameter storage cost increases to a factor $N+1$ for $N$ superclasses compared to using a single network, with finetuning, delta and quantization aware training techniques this can be reduced to $0.2N+1$. 
Another advantage of this efficient implementation is that the memory cost on the GPU during inference is equivalent to using only one network. The reason is we initiate each subclass-level network through addition of small parameter variations (deltas) to the superclass-level network. (iii) Finally, our framework promises to be more scalable and generalizable than the common alternative of simply scaling up a vanilla network in size, since very large networks often suffer from overfitting and gradient vanishing.
\end{abstract}

\section{Motivation}
DNNs have led to a series of breakthroughs for image classification \citep{krizhevsky2012imagenet,lecun1989backpropagation}. To improve  classification accuracy, researchers proposed different kinds of strategies, including designing better network structures (e.g., ResNet \cite{He_2016_CVPR} and DenseNet \cite{huang2017densely}), designing better optimizers (e.g., Adam optimizer \cite{kingma2014adam} and RMSprop optimizer \cite{tieleman2012neural}), applying better loss formulation (e.g., contrastive loss \cite{khosla2020supervised}), and using large-scale pretraining \citep{mahajan2018exploring}. 

\begin{figure}[ht]
	\begin{center}
		\includegraphics[width=\linewidth]{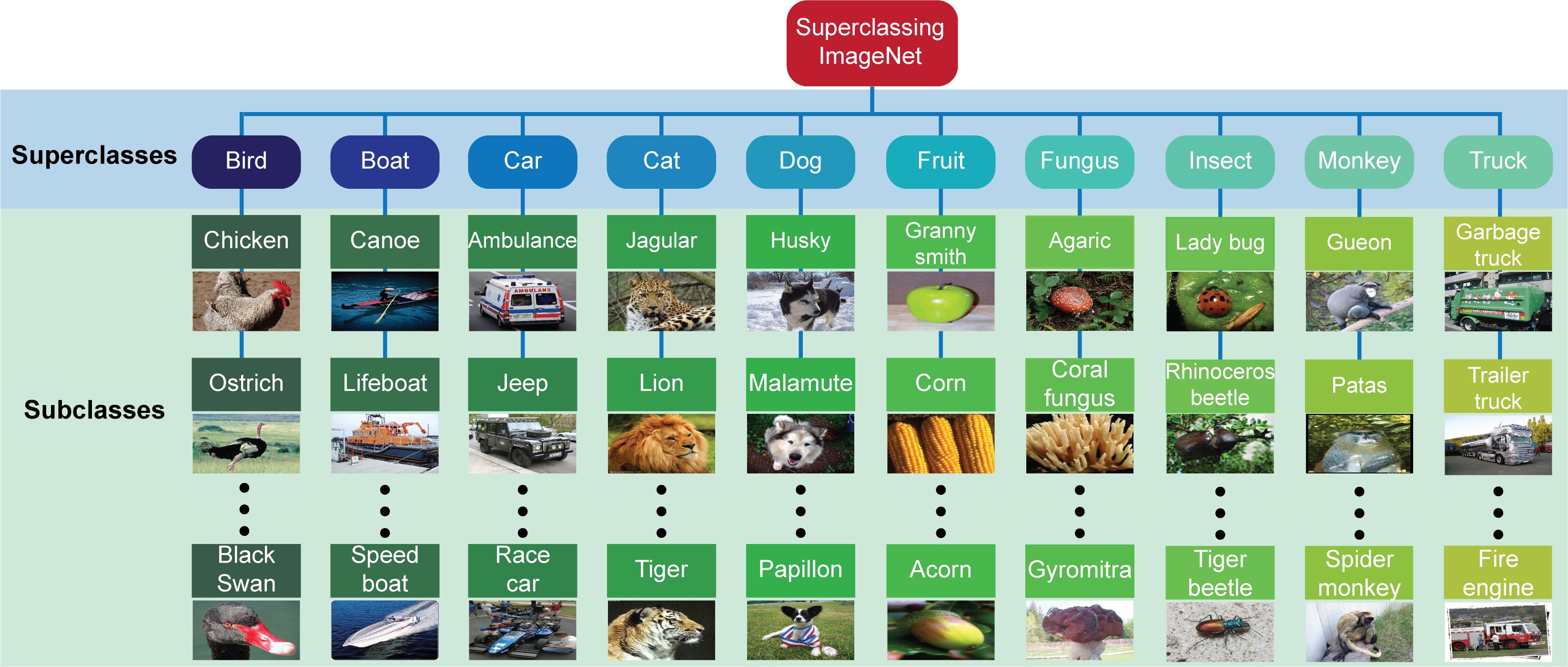}
		\caption{Superclassing ImageNet dataset. Superclassing ImageNet dataset is a subset of the ImageNet dataset. It contains broad classes which each subsume several of the original ImageNet classes. It is consisted of 10 superclasses ~--- Bird, Boat, Car, Cat, Dog, Fruit, Fungus, Insect, Monkey and Truck superclasses. Each superclass contains several subclasses, for example, the Bird superclass is consisted of Chicken, Ostrich, Black Swan subclasses and etc.}
		\label{fig:taxonomy}
	\end{center}
\end{figure}

However, few researchers have explored the patterns of misclassifications by  current DNNs. For instance, Dodge {\em et al.} \citep{dodge2017study} found that DNNs performance is much lower
than human performance on distorted images.  Understanding these patterns is particularly important because it could guide us to design more robust DNNs that make fewer mistakes and generalize better.

In addition, while in early days the answer to improving performance was often to build ever larger networks, this approach has more recently been shown to not scale well because of (i) overfitting  \citep{salman2019overfitting,srivastava2014dropout,lawrence1997lessons} and (ii) gradient vanishing \citep{hochreiter2001gradient} ~--- (i) When there is too much overfitting, this results in poor generalization in the test set. For example, Laurence {\em et al.} (\cite{lawrence1997lessons}, in their Fig.2) showed that, generally, as the network size increases, training error decreases but this ultimately leads to overfitting. As a consequence, test errors exhibit a "U" shaped behavior as a function of network size. (ii) The gradient vanishing problem limits the number of parameters and layers for scaling up the size of networks. Residual networks \cite{He_2016_CVPR} are famous for easing the training of networks that are substantially deeper. They create a gradient highway for reducing the effect of gradient vanishing. However, residual networks still cannot avoid the problem entirely. For example, in  Tab.~6 of the original residual network paper \cite{He_2016_CVPR},  performance of a residual network with 1202 layers is lower than that of residual networks with 32, 44, 56, 110 layers. We find similar scaling issues as detailed in our Results. This prompts us to explore alternatives for boosting performance with more parameters rather than  naively expanding the size of a single network.

\section{Observations and General Framework}

Often, in image classification research, datasets with the richness of ImageNet \cite{deng2009imagenet} are desired. However, the richness of the datasets imposes extraordinary difficulties for researchers to visually find any useful pattern of errors. For example, there are 1000 classes in ImageNet. It is impractical to find any useful pattern in a 1000 by 1000 confusion matrix. To smooth the path towards finding these patterns, without adding complexity to the ImageNet dataset, we “superclass” ImageNet ~--- \emph {"Superclassing ImageNet dataset"}. Superclassing ImageNet dataset is a hierarchical subset of ImageNet that contains broad superclasses (derived from Wordnet, see Sec.~\ref{Superclassing_ImageNet_datase}) which each subsume several of the original ImageNet classes (Fig.~\ref{fig:taxonomy}). Our dataset contains 10 superclasses and each superclass contains 7 to 116 subclasses, for a total of 253 subclasses. Thus, by adopting this hierarchical subdivision, we initially only navigate through 10 superclasses instead of the 1000 classes in the original ImageNet dataset. We found useful patterns in a 10 by 10 confusion matrix (Fig.~\ref{fig:confusion_matrix}) and in the summary of classification accuracies for all subclasses within each superclass (Fig.~\ref{fig:accuracy}). We highlight two observations from training convolutional neural networks on this dataset:

\begin{figure}[ht]
	\begin{center}
		\includegraphics[width=0.5\linewidth]{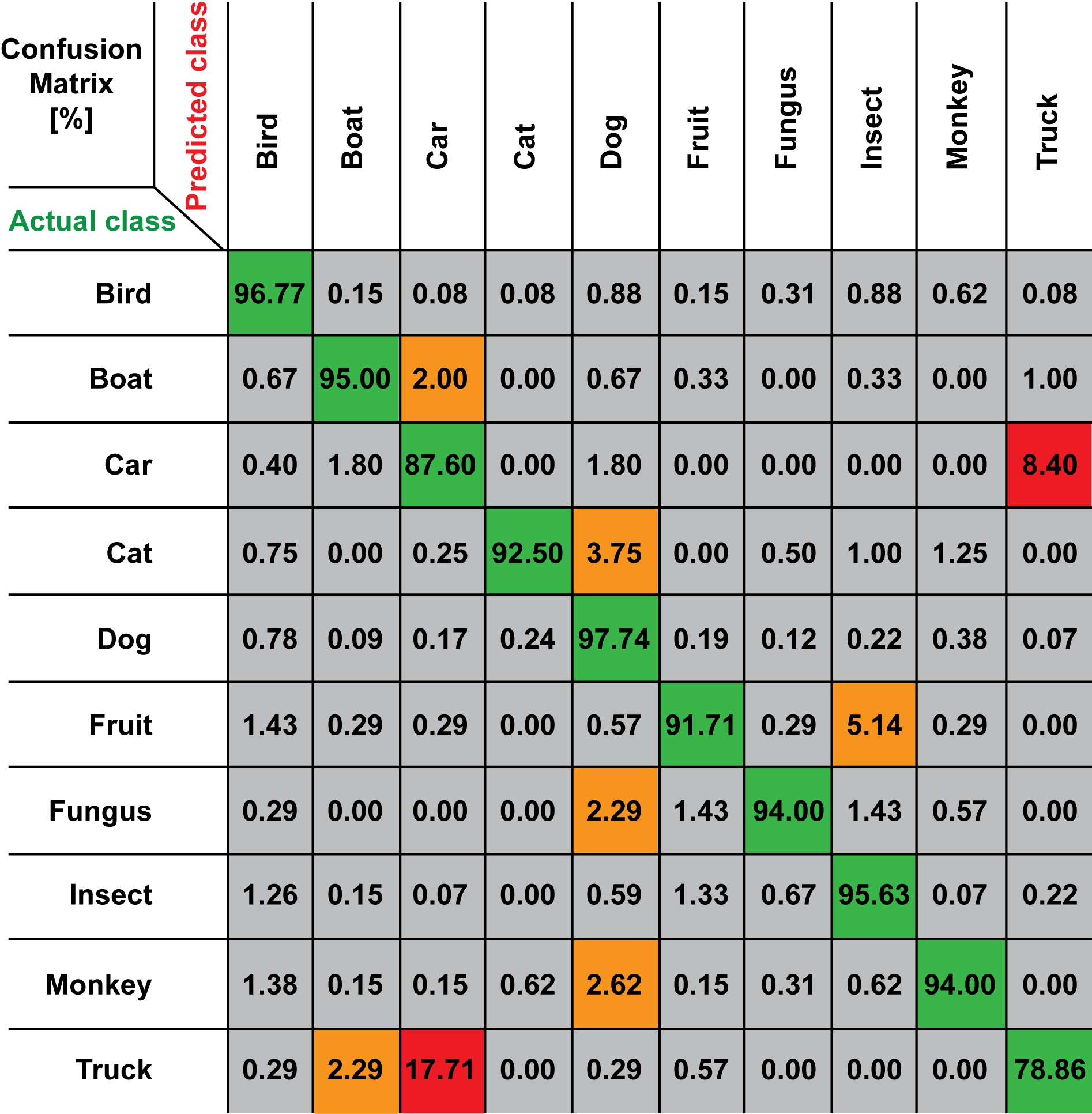}
		\caption{Confusion matrix for inter-superclasses prediction. Correct classification (green), mild wrong classification (orange), severe wrong classification (red). The convolutional neural network (ResNet-18) barely make inter-superclasses misclassifications as the averaged correct prediction performance is 96.52\%. }
		\label{fig:confusion_matrix}
	\end{center}
\end{figure}

{\bf (i) Misclassifications are seldom across superclasses  (Fig.~\ref{fig:confusion_matrix}), but mainly intra-superclass.}

We trained a ResNet-18 neural network (pretrained on ImageNet) \cite{He_2016_CVPR} on all images from all superclasses of the Superclassing ImageNet dataset to make a 10-way superclass prediction. We named it the {\bf\emph { Superclass Network}}. The high values on the diagonal of the confusion matrix in Fig.~\ref{fig:confusion_matrix} mean that the Superclass Network barely makes any inter-superclass mistakes, as the averaged correct prediction performance is 96.52\%.  For example, the probability of recognizing a bird image as belonging to the Bird superclass is 96.77\%, while the probability of recognizing a bird image as belonging to the Car superclass is only 0.08\%. In addition, for a network that is still trained on all images from all superclasses, but for which the goal is to perform a 253-way classification directly among all subclasses, the averaged correct prediction performance decreases to 71.18\%. This lower performance suggests that most misclassifications are in intra-superclass predictions.


{\bf (ii) Ensemble networks trained each only on subclasses of a given superclass (upperbound) perform better than the same network trained on all subclasses of all superclasses (lowerbound; Fig.~\ref{fig:accuracy}).} 

\label{upperbound}
In more details:

1. \emph{Upperbound (with a superclass oracle)}: We trained a ResNet-18 neural network (pretrained on ImageNet) on images from a particular superclass in the superclassing ImageNet dataset. We named it a {\bf\emph { Subclass Network}}. There are ten superclasses in the superclassing ImageNet dataset, so we have ten Subclass Networks. For a test image, with a superclass oracle, we select the right Subclass Network and only infer the subclasses inside the superclass it belongs to (upperbound). For example, a Subclass Network for the Dog superclass is trained only on dog images. It  would only predict the 116 dog subclasses (e.g., Husky, Malamute, Papillon, etc). The correct prediction performance is 75.07\% when averaged across all 10 superclasses.

2.\emph{ Lowerbound (without a superclass oracle)}: we trained a ResNet-18 neural network (pretrained on ImageNet) on all images from all subclasses in the Superclassing ImageNet dataset. For a test image, the network would infer the subclass among all 253 subclasses in all superclasses, without any prior knowledge of which superclass the test image belongs to (lowerbound). For exposition purposes, in Fig.~\ref{fig:accuracy}), the results are shown for all subclasses within each superclass. The averaged performance across all 253 subclasses is 71.18\%. The performance gap can be quantified as: upperbound is 5.47\% better than lowerbound.

\begin{figure}[h]
	\begin{center}
		\includegraphics[width=0.6\linewidth]{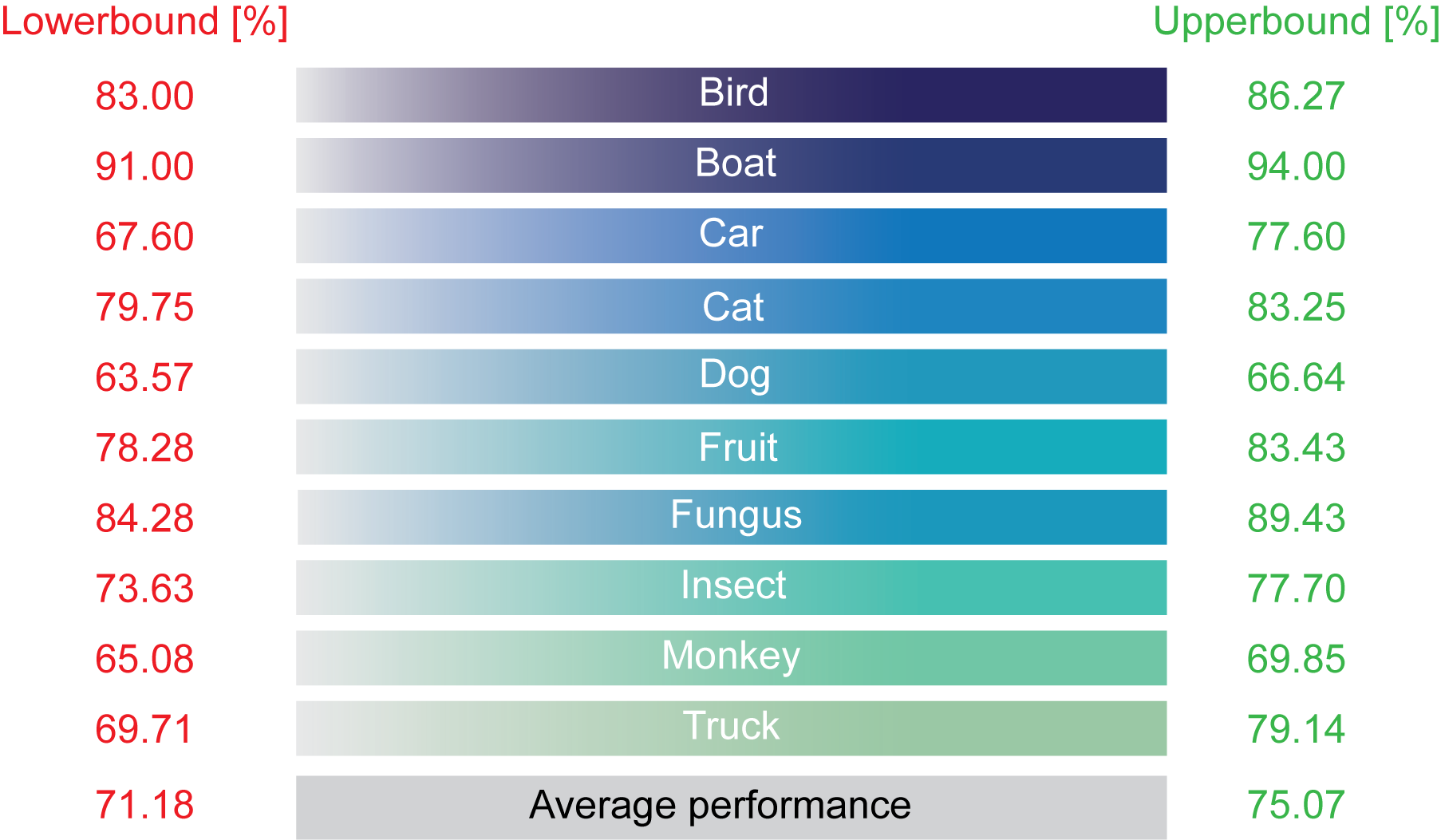}
		\caption{Performance gap between classifying an image for subclasses inside the superclass it belongs to (upperbound, green, with a superclass oracle, the averaged correct prediction performance is 75.07\%) and classifying an image for all subclasses from all superclasses (lowerbound, without a superclass oracle, red, the averaged correct prediction performance is 71.18\%). Upperbound is 5.47\% better than lowerbound.}
		\label{fig:accuracy}
	\end{center}
\end{figure}



In this paper, our goal is to design a framework such that its classification performance can \emph{ approach the upperbound performance without requiring a superclass oracle}. Here, we propose a  \emph{ two-stage Super-Sub framework} (Fig.~\ref{fig:two_stage_structure_vertical}) to reduce the performance gap. In details, in the first stage, we use a Superclass Network to decide the superclass of the test image. In the second stage, we choose the corresponding Subclass Network according to the superclass decision made in the first stage. This Subclass Network then outputs the final fine-grained subclass of the test image. More generally, our two-stage Super-Sub framework enables a neural network system to \emph{automatically adjust its behavior according to different inputs} \citep{han2021dynamic}, here achieved through the aforementioned hierarchical two-step inference. From our first observation, the Superclass Network barely makes inter-superclass mistakes, so the role of the \emph{Superclass Network can be approximated as a superclass oracle}, helping to choose the correct Subclass Network for subclass classification. As a result, our  two-stage Super-Sub framework is \emph{3.30\% better} than the lowerbound on the new Superclassing ImageNet dataset. 
Nonetheless, our vanilla implementation of the framework incurs a large cost in both parameter storage and inference runtime memory on the GPU (a factor of $N+1$ for $N$ superclasses compared to a single network). To improve this, we propose a method that keeps the same inference runtime memory as only one network, and that reduces the parameter storage cost to $0.2N+1$ based on finetuning \citep{tajbakhsh2016convolutional}, delta and quantization aware training (QAT) \citep{Quantization_aware_training} techniques, all while preserving the same accuracy boost. Most importantly, our framework promises to be more scalable and generalizable than the common alternative of simply scaling up a vanilla network in size since very large networks suffer from overfitting and gradient vanishing. We demonstrate our framwork's capacity for scalability (see Results  Sec.~\ref{results_label}) by increasing the representational capacity starting from an already very large network, i.e. Resnet-152.

\section{Our Approach}

\subsection{The Superclassing ImageNet dataset}
\begin{table}[h]
  \caption{Superclassing ImageNet dataset: For each row, we present the name of each Superclass and its contained number of subclasses, number of training images and number of test images. } 
  \label{Tab:Superclasses}
  \centering
  \begin{tabular}{cccc}
    \toprule
    \makecell {Superclass Name}     & \makecell { \# of Subclasses}     & \makecell {\# of  Training Images } & \makecell {\# of Test  Images } \\
    \midrule
    Bird  & 52  & 67,441 &  2,600   \\
    Boat     & 6 & 7,706 &  300     \\
    Car     & 10 &  13,000  & 500 \\
    Cat     & 8 &  10,400  & 400 \\
    Dog     & 116 &  145,273  & 5,800 \\\
    Fruit    & 7 &  9,100  & 350 \\
    Fungus    & 7 &  9,100  & 350 \\
    Insect    & 27 &  35,100  & 1,350 \\
    Monkey    & 13 &  16,900  & 650 \\
    Truck    & 7 &  8,959  & 350 \\
    \bottomrule
  \end{tabular}
 \end{table}
\label{Superclassing_ImageNet_datase}

\emph{Superclassing ImageNet dataset} (Fig.~\ref{fig:taxonomy}) is a new dataset that contains broad classes which each subsume several of the original ImageNet classes. It is a subset of the ImageNet Dataset that contains 10 superclasses with 253 subclasses. It contains 311,279 training images and 12,650 test images.

{\bf{Making custom dataset:}} We leverage the WordNet hierarchy according to which ImageNet is organized \citep{ImageNet_description} to create the Superclassing ImageNet dataset with 10 superclasses (Tab.~\ref{Tab:Superclasses}):


\begin{figure}[hbt!]
	\begin{center}
		\includegraphics[width=0.9\linewidth]{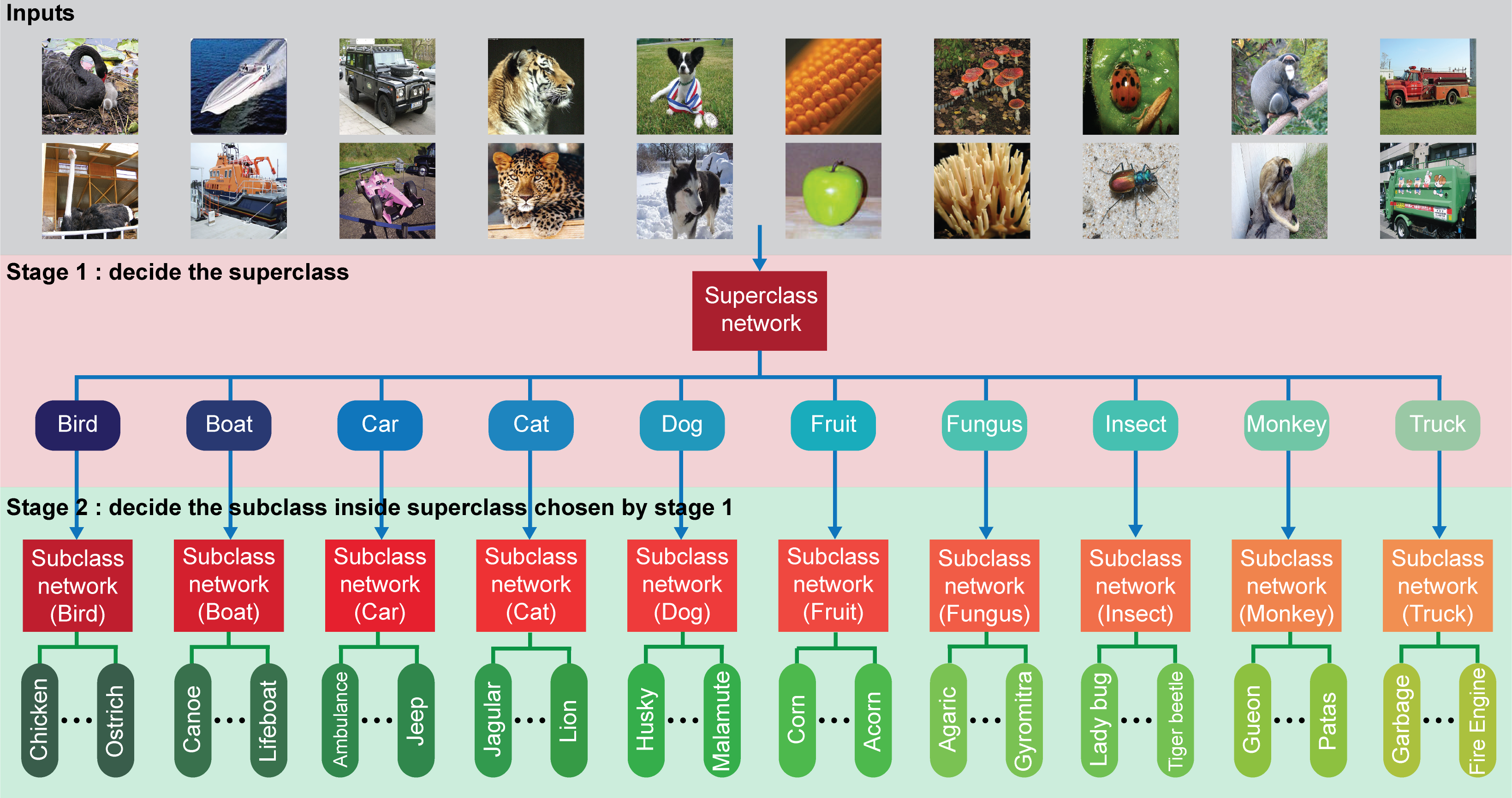}
		\caption{Two-stage Super-Sub framework: On stage 1, a Superclass Network is trained to decide the superclass of a test image. On stage 2, we choose the corresponding Subclass Network to decide the subclass inside the decided superclass in stage 1.}
		\label{fig:two_stage_structure_vertical}
	\end{center}
\end{figure}

\subsection{Super-Sub framework}

Our goal is to approach the upperbound performance without having a superclass oracle. We proposed a two-stage Super-Sub framework to achieve this goal (Fig.~\ref{fig:two_stage_structure_vertical}):  In the first stage, a Superclass Network is trained to decide the superclass of a test image. Since the Superclass Network barely makes inter-superclass mistakes, it can be approximated as a superclass oracle. The approximated superclass oracle can help us select the right Subclass Network for the second stage. In the second stage, we use the right Subclass Network to decide the subclasses inside the superclass decided in the first stage. 

\subsubsection{Vanilla Network Architectures and Vanilla Inference Rules}

\label{Vanilla}

{\bf Superclass Network:} The Superclass Network decides the superclass of a test image (e.g., Bird, Boat, Cat, etc.)  We trained a ResNet-18 neural network on all images from all superclasses of Superclassing ImageNet dataset. Formally, we trained the Superclass Network by minimizing the loss function as: 
\begin{equation}
    \begin{aligned} \mathop{\min}_{ \,\mathbf{\theta}_{super}}\end{aligned} \, \,  L\, (f_{super}\, (\mathbf{x}_{all},\, \mathbf{y}_{super}; \, \mathbf{\theta}_{super}))
    \label{Eqn:Superclass_network}
\end{equation}
where $f_{super}$ and $L$ are the Superclass Network and cross entropy loss. $\mathbf {x}_{all}$ and $\mathbf{y}_{super}$ are the image inputs from all superclasses and its superclass label. $\mathbf{\theta}_{super}$ is the parameters of the Superclass Network. The learning rate is 0.01. The  training epochs are 200.

{\bf Vanilla Subclass Networks:} Each Subclass Network decides the subclasses within the corresponding superclass.  Each Vanilla Subclass Network is a ResNet-18. We trained each Subclass Network on images from each particular superclass of our Superclassing ImageNet Dataset. Each Subclass Network would only infer in the range of the subclasses for the superclass it corresponds to. If there are $N$ superclasses, there are $N$ Subclass Networks.  For example, a Subclass Network for Bird superclass is  trained on bird images only. It  would only predict the subclasses (e.g., Chicken, Ostrich , Black Swan, etc.) within the Bird superclass.  Formally, we trained the Subclass Networks by minimizing the loss function as:
\begin{equation}
    \begin{aligned} \mathop{\min}_{ \,\mathbf{\theta}^i_{sub}}\end{aligned} \, \,  L\, (f^i_{sub}\, (\mathbf{x}^i_{sub},\, \mathbf{y}^i_{sub};\, \mathbf{\theta}^i_{sub})), \,\,\,\,\,\,\,\,\,\,\, \forall \, i \in [Bird,\, Boat,\, Car, \, ...\, Truck]
    \label{Eqn:subclass_network}
\end{equation}
where $i$ is an iterator that indicates the current superclass.  $f^i_{sub}$ is the Subclass Network corresponding to superclass $i$. $L$ is the cross entropy loss. $\mathbf{x}^i_{sub}$ and $\mathbf{y}^i_{sub}$ are the input images from superclass $i$ and its corresponding subclass labels. $\mathbf{\theta}^i_{sub}$ is the parameters of the Subclass Network for superclass $i$. For our experiments, we use a learning rate of 0.01 with 100 epochs. The settings of all hyper-parameters are the same for all subclass networks.

{\bf Vanilla Inference Rules:} 
Formally, we define the vanilla inference rules as following:

\begin{algorithm}[h]
\small
\caption{Vanilla Inference rules }
\label{alg:Vallina_inference}

\begin{algorithmic}
\STATE {Loading the Superclass Network and all Subclass Networks to GPU}
\STATE {For an image input $x$}
\STATE {\bf On the first stage:}
\STATE {\quad Deciding the superclass: $i \gets Superclass\_Network (x)$}
\STATE {\bf On the second stage:}
\STATE {\quad Choosing the corresponding $Subclass\_Network^i$ according to the $i$ in the first stage}
\STATE {\quad Deciding the subclass: $y^i_{sub} \gets Subclass\_Network^i(x)$ }
\end{algorithmic}
\end{algorithm}

\subsubsection{ Efficient Network Architectures and Efficient Inference Rules}
\label{Efficient_Network}

\begin{figure}[h]
	\begin{center}
		\includegraphics[width=0.9\linewidth]{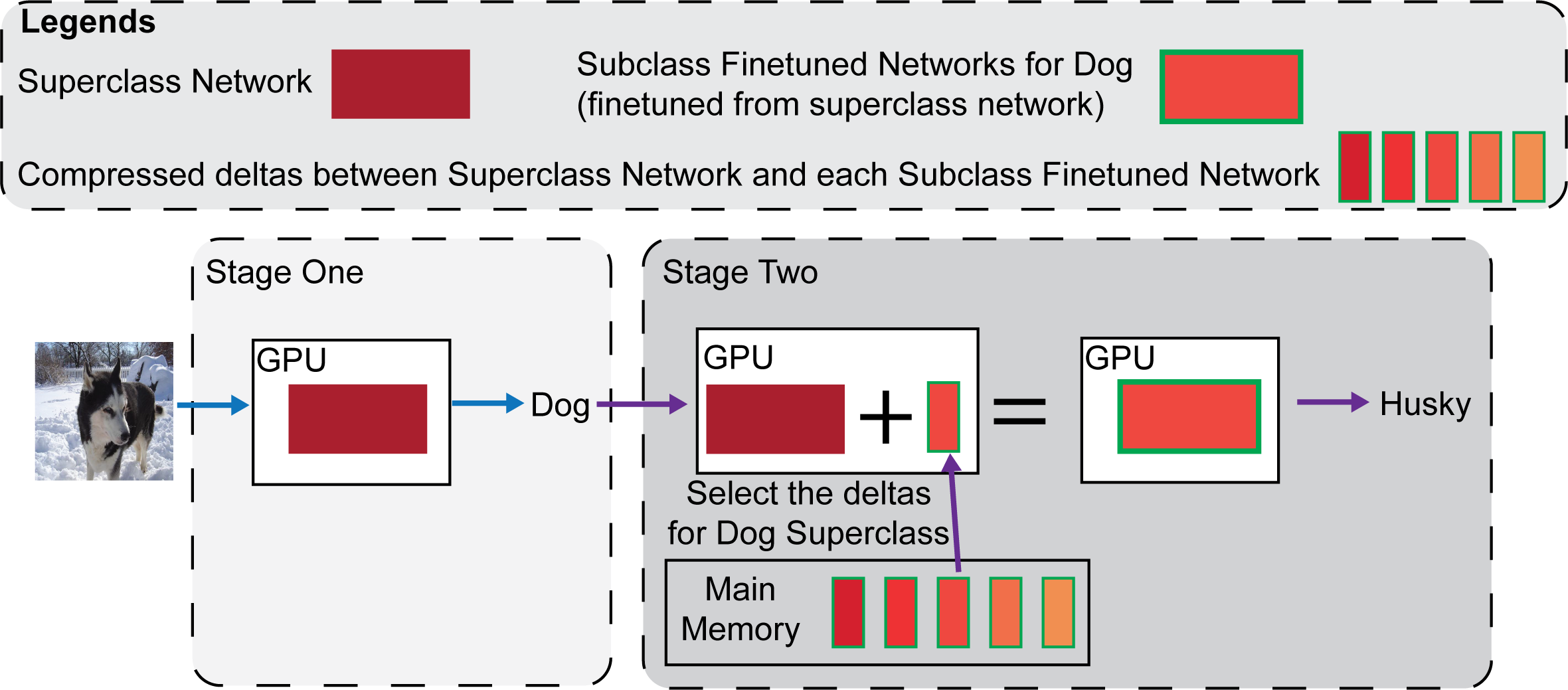}
		\caption{Efficient implementation of Super-Sub framework: one network at a time for inference. The compressed deltas (20\% of its original size) between Superclass Network and subclass finetuned networks are stored inside the main memory. The subclass finetuned networks are reconstructed by adding the corresponding deltas to the parameters of the Superclass Network. This approach is less time consuming because the compress deltas incur less I/O cost.}
		\label{fig:gpu_disk}
	\end{center}
\end{figure}

In this subsection, our goal is to implement the Super-sub framework efficiently with particular focus on deployment of deep learning systems on edge devices. It is costly to design an edge device that is capable of loading all networks to its GPU as required by our vanilla implementation. Memory-wise, it is cheaper to load one network at a time to the GPU. However, on edge devices switching between different networks incurs  expensive (in terms of time) input/output (I/O) operations between GPU and main memory. Since at inference time we need to keep switching between the super-class network and sub-network, the cost of the I/O operations would become expensive. Here, we propose an efficient implementation (Fig.~\ref{fig:gpu_disk}) that trades-off the expensive I/O operations to likely cheaper add operations.


The problem domains of Subclass Networks are subsets of the problem domain of the Superclass Network. For example, recognizing an image of a Husky in the dog superclass is a subproblem of recognizing an image of dog in various superclasses.  Since it is a domain reduction problem \citep{tawarmalani2013convexification}, Subclass Networks can be created by finetuning from the Superclass Network efficiently (Subclass Finetuned Networks). We found that the deltas between the parameters of the Superclass Network and each Subclass Finetuned Network are small and because of this, we are able to greatly compress them. Thus, in our efficient implementation, to recreate a Subclass Network, we load the compressed deltas (reducing the I/O operations to 20\%), and add them in the GPU to the parameters of the Superclass Network (Fig.~\ref{fig:gpu_disk}). 

{\bf Superclass Network}: The network architecture and training of the Superclass Network in the efficient implementation is the same as the Vanilla Superclass Network described in  Sec.~\ref{Vanilla}.

{\bf Subclass Finetuned Network}: Instead of training each Subclass Network from scratch for each superclass, we created it by finetuning the Superclass network on the images from each particular superclass of the superclassing ImageNet dataset, by minimizing the loss function:

\begin{equation}
\resizebox{0.945\textwidth}{!}
    {$f^i_{sub\_finetuned}\, (\mathbf{x}^i_{sub},\mathbf{y}^i_{sub};\mathbf{\theta}^i_{sub}) \gets \begin{aligned} \mathop{\min}_{ \,\mathbf{\theta}_{super}}\end{aligned} \, \,  L\, (f_{super}\, (\mathbf{x}^i_{sub},\mathbf{y}^i_{sub}; \mathbf{\theta}_{super})), \,\,\,\, \forall \, i \in [Bird,\, Boat,\, ...\, Truck]$}
    \label{Eqn:finetuning_subclass_network}
\end{equation}
where, $f^i_{sub\_finetuned}$ is the Subclass Finetuned Network for superclass $i$, other notations are the same as in Eqn.~\ref{Eqn:Superclass_network} and Eqn.~\ref{Eqn:subclass_network}. The hyper-parameters are the same in the Vanilla Subclass Networks.

{\bf Deltas}: The deltas for a particular superclass are created with the subtraction of the parameters of the Superclass Network from that of the corresponding Subclass Finetuned Network:
\begin{equation}
    deltas^i = \mathbf{\theta}^i_{sub} - \mathbf{\theta}_{super} 
    \label{Eqn:deltas}
\end{equation}
where $deltas^i$ corresponds to the deltas for superclass $i$. other notations are as in Eqn.~\ref{Eqn:Superclass_network} and Eqn.~\ref{Eqn:subclass_network}. $deltas^i$ is calculated through mini-batch gradient descent via Eqn.~\ref{Eqn:finetuning_subclass_network}. Since these gradients are in half-precision (16 bits), $deltas^i$ is represented in half-precision (50\% of the size to store the Subclass Finetuned network in full-precision 32 bits). Then, we use 7-Zip to further compress $deltas^i$ (reduce to 44\% of the size to store the Subclasses Finetuned network in full-precision) and store them. In the following, we show how to further reduce the compressed delta size to a final 20\% with Quantization Aware Training techniques. For inference, to recreate the Subclass Finetuned Network, we can load the corresponding compressed deltas to GPU and add them to the Superclass Network.

{\bf Efficient Inference Rules}: Formally, we defined the efficient inference rules as:
\label{efficient_interferen_rules}

\begin{algorithm}[h]
\small
\caption{Efficient Inference rules }
\label{alg:Efficient_inference}

\begin{algorithmic}
\STATE {Loading the Superclass Network to GPU}
\STATE {For an image input $x$}
\STATE {\bf On the first stage:}
\STATE {\quad Deciding the superclass: $i \gets Superclass\_Network (x)$}
\STATE {\bf On the second stage:}
\STATE {\quad Loading the compressed $deltas^i$ according to the superclass $i$}
\STATE {\quad Recreating the Subclass Finetuned Network for superclass $i$ through:     $\mathbf{\theta}^i_{sub} \gets deltas^i + \mathbf{\theta}^i_{super}$ }
\STATE {\quad Deciding the subclass: $y^i_{sub} \gets Subclass\_Finetuned\_Network^i(x)$ }
\end{algorithmic}
\end{algorithm}

{\bf Quantization Aware Training (QAT):} To further reduce the compressed size of the deltas, we used Quantization aware training (QAT) \citep{Quantization_aware_training}, available in the Pytorch Framework, to finetune the Superclass Network (See Supplementary).  With  QAT, the  values are limited to 256 possibilities in 8 bit quantization. Thus, with less variability, the compressed size of the deltas is further reduced to 20\% compared to storing the original quantized Subclass Network.  

In summary, in our efficient implementation there is always only one network running on the GPU. Furthermore, the extra main memory cost for each Subclass Network is reduced to around 20\% of its original size. 


%

\section{Experiments}
We evaluated our Vanilla and Efficient implementations on the Superclassing ImageNet Dataset. We report  classification accuracy on each superclass and  average accuracy across all superclasses. In addition, we report the compression ratio for storing the deltas used in the efficient implementation. 

To our best knowledge, we are the first to leverage the patterns of misclassified images to improve the scalability and generalizability of DNNs. Because our method is in a very novel direction, 
we did not find directly applicable benchmarks to compare to. Instead, we propose the lowerbound and upperbound defined in Sec.~\ref{upperbound} to act as natural references to our framework's performance. 









\section{Results and Analysis}

\begin{figure}[htb]
	\begin{center}
		\includegraphics[width=0.85\linewidth]{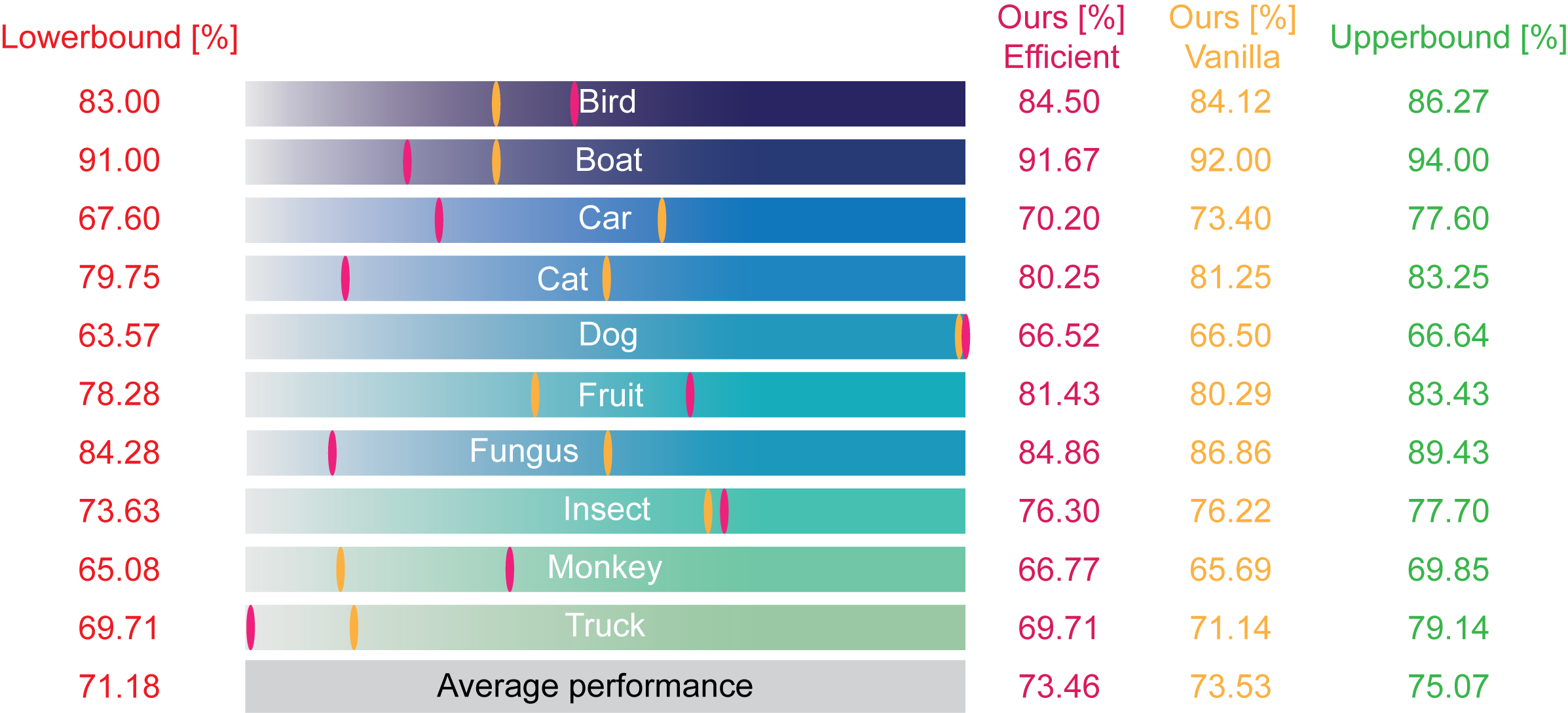}
		\caption{The performances of Efficient (rose-red) and Vanilla (yellow) Implementations of the Super-Sub framework. Both Efficient and Vanilla Implementations reduced the performance gap between upperbound (green) and lowerbound (red). The rose-red and yellow eclipses represent the relative position of efficient and vanilla implementations' performances to the upperbound and lowerbound.     The definitions of upperbound and lowerbound are the same as that of Fig.~\ref{fig:accuracy}.  }
		\label{fig:accuracy_attention}
	\end{center}
\end{figure}

{\bf The Super-Sub framework improved overall classification performance without a superclass oracle:}  On average, with a RestNet-18 backbone, our Efficient (Vanilla) implementation is 3.20\% (3.30\%) better than the lowerbound (Fig.~\ref{fig:accuracy_attention}) and approaches the upperbound. The performances of Efficient and Vanilla implementations are almost equivalent. With a RestNet-152 backbone,  our Vanilla implementation is 8.00\% better than the lowerbound. 

\begin{figure}[ht]
	\begin{center}
		\includegraphics[width=0.85\linewidth]{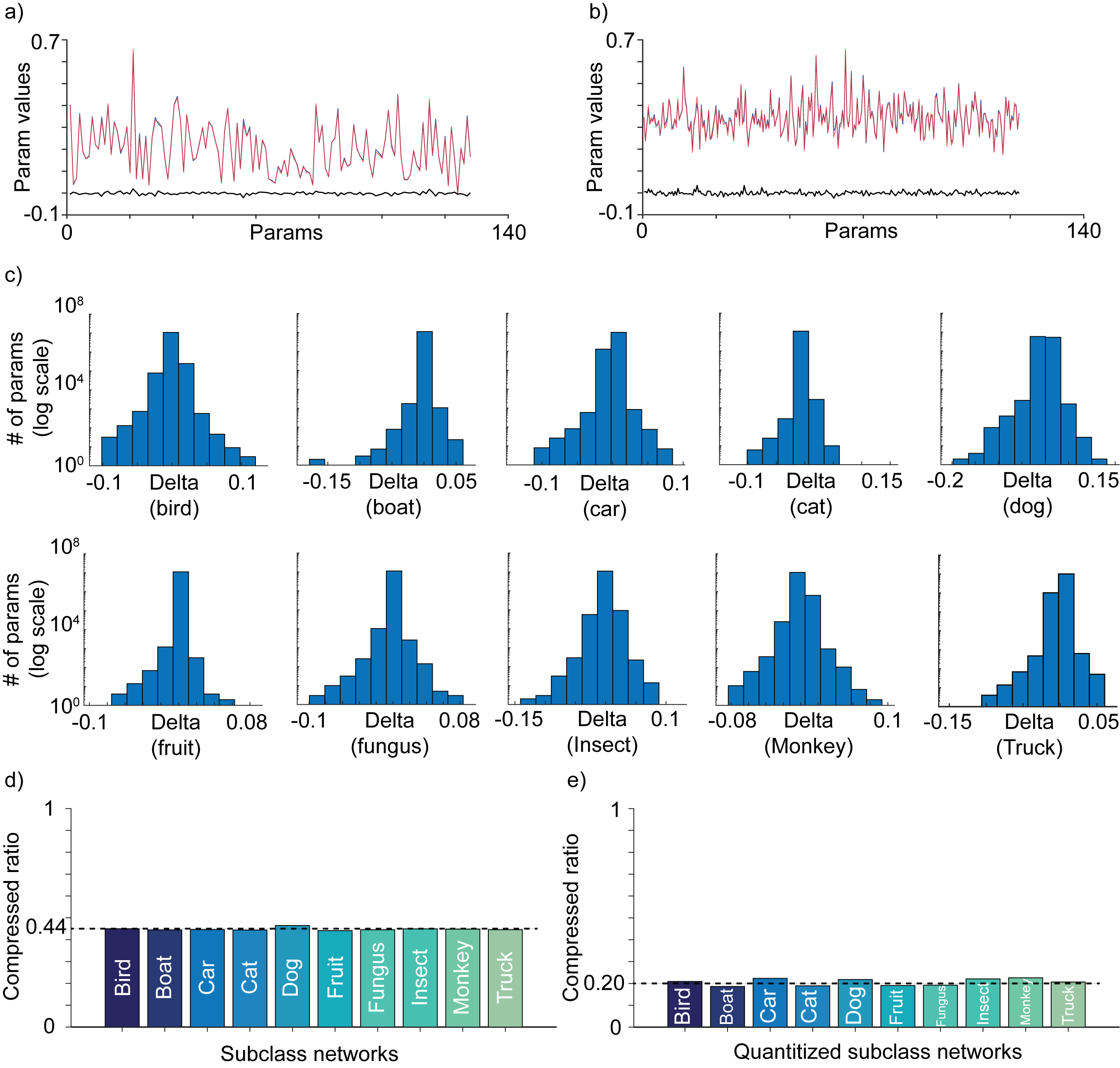}
		\caption{a-b) Visualization of weights of a convolutional layer and a batch normalization layer in a Superclass Network (red) and Subclass Finetuned Network (for fruit superclass, blue). The black curve is the difference between the two networks. c) Various histograms of deltas between Superclass network and Subclass Finetuned Network for each superclass.  For d-e) The dash line is the averaged compressed ratio across all superclasses. d) Compressed ratio for Subclass Networks without using QAT.  e) Compressed ratio for Subclass Networks with QAT.}
		\label{fig:delta_log}
	\end{center}
\end{figure}

{\bf Deltas are small and can be greatly compressed:} We found that the deltas of parameters between the Superclass Network and Subclass Finetuned Networks are small (Fig.~\ref{fig:delta_log}.a,b; black curve  fluctuates around 0). The histograms of deltas for different superclasses further confirmed this (Fig.~\ref{fig:delta_log}.c). In addition, we found we can compress the deltas because their distributions are highly concentrated around 0 (Fig.~\ref{fig:delta_log}.c). The average compression ratio is 0.44 without using QAT (Fig.~\ref{fig:delta_log}.d) and is 0.20 with QAT (Fig.~\ref{fig:delta_log}.e). This means that the GPU only needs to load 20\% of the size of the original quantized Subclass Finetuned Network to recreate it from an already loaded Superclass Network. 

 \label{results_label}
 
{\bf Super-sub framework promises to be more scalable and generalizable:} To demonstrate this, we pick a starting point of a large network (ResNet-152) as a backbone of our Super-Sub framework. Resnet-152, in terms of memory size, marks a threshold after which gradient vanishing and overfitting can start occurring, prohibiting further layer increases within the traditional ResNet regime. This is often referred as the "U" shaped test error performance as a function of network size. We show that if we scale up Resnet-152 using our super-sub framework, which incurs in adding 440 MB of memory, we observe a huge performance boost from 78.43\% to 84.63\% in accuracy. In contrast, if Resnet-152 is scaled up naively just by adding layers, which here would be approximately equal to expanding to a Resnet-455 (Super-Sub framework with a Resnet-152 backbone $\approx$ Resnet-455 in terms of memory), we observe a stark performance degradation, with the accuracy lowering to 68.01\%.


\section{Discussion}
\label{discussion}

In this paper, we studied the patterns of misclassified ImageNet images and leveraged these patterns to improve the scalability and generalizability of a DNN. By training a neural network on "Superclassing ImageNet Dataset", we found that: (i) Misclassifications are rarely inter-superclasses  but mainly  intra-superclass. (ii) Ensemble networks trained each only on subclasses of a given superclass perform better than the same network trained on all subclasses of all superclasses. Hence, we proposed  the Super-Sub framework and demonstrated that: (i) The Super-Sub framework reduced the performance gap without having a superclass oracle, (ii)  We leverage \citep{Quantization_aware_training} techniques to create an efficient implementation with low parameter and memory costs. (iii) We demonstrated that our Super-Sub framework promises to be more scalable and generalizable.

To date, scaling up the size and depth of DNN quickly becomes prohibitive due to the emergence of gradient vanishing \citep{hochreiter2001gradient} and overfitting issues. Our two-stage super-sub framework provides a new and efficient way of scaling up a DNN beyond the current limitations. Our framework subdivides the overall classification problem domain into separate smaller sub-domains that are semantically grouped. Each image goes first through a high-level classification by a Superclass Network which then routes it to the appropriate Subclass Network specialized for the sub-domain fine-grained classification. In our approach, each sub-domain is only a subset of the original overall classification problem. Hence, each can be expressed through a smaller model with less parameters and therefore less overfitting when compared to using a larger model to tackle directly the original problem. As previously discussed, naively increasing the size and depth within one DNN is not scalable because of gradient vanishing. Here, our model addresses this by increasing the number of parameters "laterally" into several separate network modules (sub-domains) which naturally confers more robustness to gradient vanishing and allows a much greater potential for upward scalability.

In sum, our approach is a more scalable and generalizable alternative to just naively expanding the size of a single network. Nonetheless, despite those excellent benefits, two-stage Super-Sub framework has some limitations: (i) Errors in first stage are unrecoverable; We could create better superclasses to maximize the Superclass Network performance using a data-driven approach. (ii) We have not yet fully implemented the efficient version on an edge device. This may require a next generation of devices that could support uncompress, weight operations, etc, if our approach gains sufficient general interest.

\clearpage
\bibliography{neurips_2021}
\bibliographystyle{unsrt}

\section{Supplementary}
\subsection{Quantization Aware Training (QAT):}
"QAT is the quantization method that typically results in the highest accuracy . With QAT, all weights and activations are 'fake quantized' during both the forward and backward passes of training: that is, float values are rounded to mimic int8 values, but all computations are still done with floating point numbers. Thus, all the weight adjustments during training are made while “aware” of the fact that the model will ultimately be quantized; after quantizing, therefore, this method will usually yield higher accuracy than either dynamic quantization or post-training static quantization."\citep{Quantization_aware_training}. 

\section*{Checklist}
\begin{enumerate}

\item For all authors...
\begin{enumerate}
  \item Do the main claims made in the abstract and introduction accurately reflect the paper's contributions and scope?
  
    Yes.
  \item Did you describe the limitations of your work?
  
    Yes, We descriped them in the discussion section.
  \item Did you discuss any potential negative societal impacts of your work?
  
    We don't know  any potential negative societal impacts of out work.
  \item Have you read the ethics review guidelines and ensured that your paper conforms to them?
    Yes, it conforms.
\end{enumerate}


\item If you ran experiments...
\begin{enumerate}
  \item Did you include the code, data, and instructions needed to reproduce the main experimental results (either in the supplemental material or as a URL)?
  
    Upon publication, we will release them.
    
  \item Did you specify all the training details (e.g., data splits, hyperparameters, how they were chosen)?
  
    Yes, we include it in the Section 2 and Section 3.
    
	\item Did you report error bars (e.g., with respect to the random seed after running experiments multiple times)?
	
    We ran the experiments multiple times and can reproduce the results.
    
	\item Did you include the total amount of compute and the type of resources used (e.g., type of GPUs, internal cluster, or cloud provider)?
	
    We used Nvidia 1080Ti and Nvidia V100 to train the model.  
\end{enumerate}



\end{enumerate}
\end{document}